# SIGNIFICANCE OF CLASSIFICATION TECHNIQUES IN PREDICTION OF LEARNING DISABILITIES

Julie M. David[1] and Kannan Balakrishnan[2]

[1]MES College, Aluva, Cochin- 683 107, India
julieeldhosem@yahoo.com
[2]Cochin University of Science & Technology, Cochin - 682 022, India
Mullayilkannan@gmail.com

## ABSTRACT

*The aim of this study is to show the importance of two classification techniques, viz. decision tree and clustering, in prediction of learning disabilities (LD) of school-age children. LDs affect about 10 percent of all children enrolled in schools. The problems of children with specific learning disabilities have been a cause of concern to parents and teachers for some time. Decision trees and clustering are powerful and popular tools used for classification and prediction in Data mining. Different rules extracted from the decision tree are used for prediction of learning disabilities. Clustering is the assignment of a set of observations into subsets, called clusters, which are useful in finding the different signs and symptoms (attributes) present in the LD affected child. In this paper, J48 algorithm is used for constructing the decision tree and K-means algorithm is used for creating the clusters. By applying these classification techniques, LD in any child can be identified.*

## 1. INTRODUCTION

Data mining is a collection of techniques for efficient automated discovery of previously unknown, valid, novel, useful and understandable patterns in large databases. Conventionally, the information that is mined is denoted as a model of the semantic structure of the datasets. The model might be utilized for prediction and categorization of new data [1]. In recent years the sizes of databases has increased rapidly. This has lead to a growing interest in the development of tools capable in the automatic extraction of knowledge from data. The term Data Mining or Knowledge Discovery in databases has been adopted for a field of research dealing with the automatic discovery of implicit information or knowledge within databases [16]. Diverse fields such as marketing, customer relationship management, engineering, medicine, crime analysis, expert prediction, web mining and mobile computing besides others utilize data mining [7].

Databases are rich with hidden information, which can be used for intelligent decision making. Classification and prediction are two forms of data analysis that can be used to extract models describing important data classes or to predict future data trends [8]. Classification is a data mining (machine learning) technique used to predict group membership for data instances. Machine learning refers to a system that has the capability to automatically learn knowledge from experience and other ways [4]. Classification predicts categorical labels whereas prediction models continuous valued functions. Classification is the task of generalizing known structure to apply to new data while clustering is the task of discovering groups and structures in the data that are in some way or another similar, without using known structures in the data.

Decision trees are supervised algorithms which recursively partition the data based on its attributes, until some stopping condition is reached [8]. This recursive partitioning, gives rise to a tree-like structure. Decision trees are white boxes as the classification rules learned by them can be easily obtained by tracing the path from the root node to each leaf node in the tree. Decision trees are very efficient even with the large volumes data. This is due to the partitioning nature of the algorithm, each time working on smaller and smaller pieces of the dataset and the fact that they usually only work with simple attribute-value data which is easy to manipulate. The Decision Tree Classifier (DTC) is one of the possible approaches to multistage decision-making. The most important feature of DTCs is their capability to break down a complex decision making process into a collection of simpler decisions, thus providing a solution, which is often easier to interpret [17].

Clustering is the one of the major data mining tasks and aims at grouping the data objects into meaningful classes or clusters such that the similarity of objects within clusters is maximized and the similarity of objects from different clusters is minimized [10]. Clustering separates data into groups whose members belong together. Each object is assigned to the group it is most similar to. Cluster analysis is a good way for quick review of data, especially if the objects are classified into many groups. Clustering does not require a prior knowledge of the groups that are formed and the members who must belong to it. Clustering is an unsupervised algorithm [6]. Clustering is often confused with classification, but there is some difference between the two. In classification the objects are assigned to pre defined classes, whereas in clustering the classes are also to be defined [11].

## 2. LEARNING DISABILITY

LD is a neurological condition that affects a child's brain and impairs his ability to carry out one or many specific tasks. These like children are neither slow nor mentally retarded. An affected child can have normal or above average intelligence. This is why a child with a learning disability is often wrongly labeled as being smart but lazy. LDs affect about 10 percent of all children enrolled in schools. The problems of children with specific learning disabilities have been a cause of concern to parents and teachers for some time. Pediatricians are often called on to diagnose specific learning disabilities in school- age children. Learning disabilities affect children both academically and socially. These may be detected only after a child begins school and faces difficulties in acquiring basic academic skills [11]. Learning disability is a general term that describes specific kinds of learning problems.

Specific learning disabilities have been recognized in some countries for much of the 20th century, in other countries only in the latter half of the century, and yet not at all in other places [11]. A learning disability can cause a person to have trouble learning and using certain skills. The skills most often affected are: reading, writing, listening, speaking, reasoning, and doing math. If a child has unexpected problems or struggling to do any one of these skills, then teachers and parents may want to investigate more. The child may need to be evaluated to see if he or she has a learning disability.

Learning disabilities are formally defined in many ways in many countries. However, they usually contain three essential elements: a discrepancy clause, an exclusion clause and an etiologic clause. The discrepancy clause states there is a significant disparity between aspects of specific functioning and general ability; the exclusion clause states the disparity is not primarily due to intellectual, physical, emotional, or environmental problems; and the etiologic clause speaks to causation involving genetic, biochemical, or neurological factors. The most frequent clause used in determining whether a child has a learning disability is the difference between areas of functioning. When a child shows a great disparity between those areas of functioning in which she or he does well and those in which considerable difficulty is experienced, this child is described as having a learning disability [12]. Learning disabilities vary from child to child. One child with LD may not have the same kind of learning problems as another child with LD. There is no "cure" for learning disabilities [14]. They are life-long. However, children with LD can be high achievers and can be taught ways to get around the learning disability. With the right help, children with LD can and do learn successfully. There is no *one sign* that shows a child has a learning disability. Experts look for a noticeable difference between how well a child does in school and how well he or she could do, given his or her intelligence or ability. There are also certain clues, most relate to elementary school tasks, because learning disabilities tend to be identified in elementary school, which may mean a child has a learning disability. A child probably won't show all of these signs, or even most of them

When a LD is suspected based on parent and/or teacher observations, a formal evaluation of the child is necessary. A parent can request this evaluation, or the school might advise it. Parental consent is needed before a child can be tested [12]. Many types of assessment tests are available. Child's age and the type of problem determines the tests that child needs. Just as there are many different types of LDs, there are a variety of tests that may be done to pinpoint the problem. A complete evaluation often begins with a physical examination and testing to rule out any visual or hearing impairment [3]. Many other professionals can be involved in the testing process.

The purpose of any evaluation for LDs is to determine child's strengths and weaknesses and to understand how he or she best learns and where they have difficulty [12]. The information gained from an evaluation is crucial for finding out how the parents and the school authorities can provide the best possible learning environment for child.

## 3. PROPOSED APPROACH

This study consists of two parts. In the former part, LD prediction is classified by using decision tree and in the latter part by clustering. J48 algorithm is used in constructing the decision tree and K-means algorithm is used in creating the clusters of LD.

A decision is a flow chart like structure, where each internal node denotes a test on an attribute, each branch of the tree represents an outcome of the test and each leaf node holds a class label [8]. The topmost node in a tree is the root node. Decision tree is a classifier in the form of a tree structure where each node is either a leaf node-indicates the value of the target attribute of examples or a decision node –specifies some test to be carried out on a single attribute-with one branch and sub tree for each possible outcome of the test[9]. Decision tree can handle high dimensional data. The learning and classification step of decision tree are simple and fast. A decision tree can be used to classify an example by starting at the root of the tree and moving through it until a leaf node, which provides the classification of the instance [17]. In this work we are using the well known and frequently used algorithm J48 for the classification of LD. To classify an unknown instance, it is routed down the tree according to the values of the attributes tested in successive nodes and when a leaf is reached, the instance is classified according to the class assigned to the leaf [17].

Clustering is a tool for data analysis, which solves classification problem. Its object is to distribute cases into groups, so that the degree of association to be strong between members of same clusters and weak between members of different clusters. This way each cluster describes in terms of data collected, the class to which its members belong. Clustering is a discovery tool. It may reveal associations and structure in data which though not previously evident .The results of cluster analysis may contribute to the definition of a formal classification scheme. Clustering helps us to find natural groups of components based on some similarity. Clustering is the assignment of a set of observations into subsets so that observations in the same cluster are similar in some sense.

Clustering is a method of unsupervised learning, and a common technique for statistical data analysis used in many fields, including machine learning, data mining, pattern recognition, image analysis and bioinformatics.

## 3.1 Classification by Decision Tree

Data mining techniques are useful for predicting and understanding the frequent signs and symptoms of behavior of LD. There are different types of learning disabilities. If we study the signs and symptoms (attributes) of LD we can easily predict which attribute is from the data sets more related to learning disability. The first task to handle learning disability is to construct a database consisting of the signs, characteristics and level of difficulties faced by those children. Data mining can be used as a tool for analyzing complex decision tables associated with the learning disabilities. Our goal is to provide concise and accurate set of diagnostic attributes, which can be implemented in a user friendly and automated fashion. After identifying the dependencies between these diagnostic attributes, rules are generated and these rules are then be used to predict learning disability. In this paper, we are using a checklist containing the same 16 most frequent signs & symptoms (attributes) generally used for the assessment of LD [13] to investigate the presence of learning disability. This checklist is a series of questions that are general indicators of learning disabilities. It is not a screening activity or an assessment, but a checklist to focus our understanding of learning disability. The list of 16 attributes used by us in LD prediction is shown in Table 1 below.

Table 1. List of Attributes

| Sl. No. | Attribute | Signs & Symptoms of LD |
|---|---|---|
| 1 | DR | Difficulty with Reading |
| 2 | DS | Difficulty with Spelling |
| 3 | DH | Difficulty with Handwriting |
| 4 | DWE | Difficulty with Written Expression |
| 5 | DBA | Difficulty with Basic Arithmetic skills |
| 6 | DHA | Difficulty with Higher Arithmetic skills |
| 7 | DA | Difficulty with Attention |
| 8 | ED | Easily Distracted |
| 9 | DM | Difficulty with Memory |
| 10 | LM | Lack of Motivation |
| 11 | DSS | Difficulty with Study Skills |
| 12 | DNS | Does Not like School |
| 13 | DLL | Difficulty Learning a Language |
| 14 | DLS | Difficulty Learning a Subject |
| 15 | STL | Slow To Learn |
| 16 | RG | Repeated a Grade |

Based on the information obtained from the checklist, a data set is generated. This is set is in the form of an information system containing cases, attributes and class. A complete information system expresses all the knowledge available about objects being studied. Decision tree induction is the learning of decisions from class labeled training tuples. Given a data set D = {$t_1$, $t_2$,.........., $t_n$} where $t_i$ = <$t_{i1}$,....., $t_{ih}$>. In our study, each tuple is represented by 16 attributes and the class is LD. Then, Decision or Classification Tree is a tree associated with D such that each internal node is labeled with attributes DR, DS, DH, DWE, etc. Each arc is labeled with predicate, which can be applied to the attribute at the parent node. Each leaf node is labeled with a class LD. The basic steps in the decision tree are building the tree by using the training data sets and applying the tree to the new data sets. Decision tree induction is the process of learning about the classification using the inductive approach [8]. During this process we create a new decision tree from the training data. This decision tree can be used for making classifications. Here we are using the J48 algorithm, which is a greedy approach in which decision trees are constructed in a top-down recursive divide and conquer manner. Most algorithms for decision tree approach are following such a top down approach. It starts with a training set of tuples and their associated class labels. The training set is recursively partitioned into smaller subsets as a tree is being built. This algorithm consists of three parameters – attribute list, attribute selection method and classification. The attribute list is a list of attributes describing the tuples. Attribute selection method specifies a heuristic procedure for selecting the attribute that best discriminate the given tuples according to the class. The procedure employs an attribute selection measure such as information gain that allows a multi-way splits. Attribute selection method determines the splitting criteria. The splitting criteria tells as which attribute to test at a node by determining the best way to separate or partition the tuples into individual classes. Here we are using the data mining tool weka for attribute selection and classification. Classification is a data mining (Machine Learning) technique, used to predict group membership from data instances [15].

### 3.1.1 Methodology used

J48 algorithm is used for classifying the Learning Disability. The procedure consists of three steps viz. (i) data partition based on cross validation test, (ii) attribute list and (iii) attribute selection method based on information gain. Cross validation approach is used for the sub sampling of datasets. In this approach, each record is used the same number of times for training and exactly once for testing. To illustrate this method, first we partition the datasets into two subsets and choose one of the subsets for training and other for testing. Then swap the roles of the subsets so that the previous training set becomes the test set and vice versa. The Information Gain Ratio for a test is defined as follows. IGR (Ex, a) = IG / IV, where IG is the Information Gain and IV is the Gain Ratio [13]. Information gain ratio biases the decision tree against considering attributes with a large number of distinct values. So it solves the drawback of information gain. The classification results are as shown under:
Correctly Classified Instances       97 Nos.   77.6 %
Incorrectly Classified Instances     28 Nos.   22.4 %
The accuracy of the decision tree is given in Table 2 below.

Table 2. Accuracy of Decision Tree

| TP Rate | FP Rate | Precision | Recall | F-Measure | ROC Area | Class |
|---------|---------|-----------|--------|-----------|----------|-------|
| 0.840 | 0.419 | 0.859 | 0.840 | 0.849 | 0.719 | N |
| 0.581 | 0.160 | 0.545 | 0.581 | 0.563 | 0.719 | Y |

The first two columns in the table denote TP Rate (True Positive Rate) and the FP Rate (False Positive Rate). TP Rate is the ratio of low weight cases predicted correctly cases to the total of positive cases. A decision tree formed based on the methodology adopted in this paper is shown in Figure 1 below.

It is easy to read a set of rules directly off a decision tree. One rule is generated for each leaf. The antecedent of the rule includes a condition for every node on the path from the root to that leaf and the consequent of the rule is the class assigned by the leaf [17]. This procedure produces rules that are unambiguous in that the order in which they are executed is irrelevant. However in general, rules that are read directly off a decision tree are far more complex than necessary and rules derived from trees are usually pruned to remove redundant tests. The rules are so popular because each rule represents an independent knowledge. New rule can added to an existing rule sets without disturbing them, whereas to add to a tree structure may require reshaping the whole tree. In this section we present a method for generating a rule set from a decision tree. In principle, every path from the root node to the leaf node of a decision tree can be expressed as a classification rule. The test conditions encountered along the path form the conjuncts of the rule antecedent, while the class label at the leaf node is assigned to the rule consequent. The expressiveness of a rule set is almost equivalent to that of a decision tree because a decision tree can be expressed by a set of mutually exclusive and exhaustive rules.

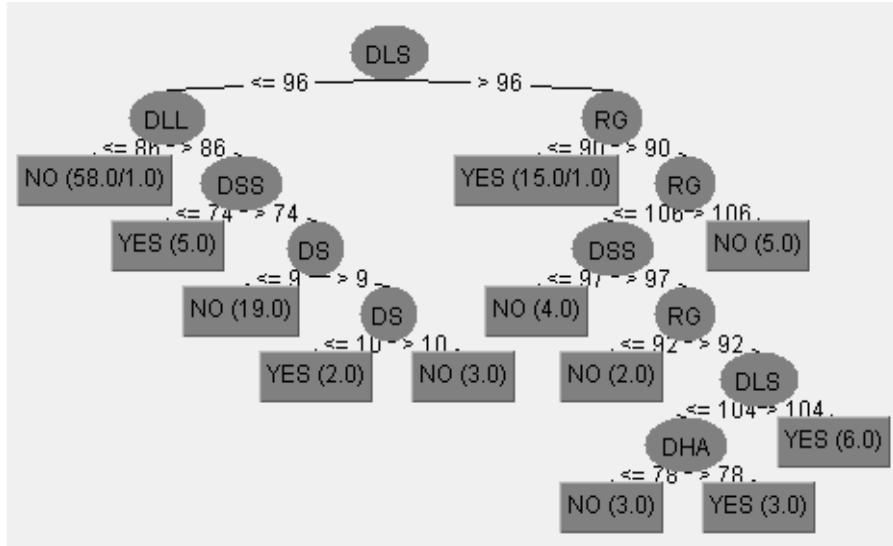

Figure 1. Decision tree

## 3.2 Classification by Clustering

We are using the data mining tool weka for clustering. The clustering algorithm K-means is used for classifying LD. In clustering algorithm, K initial pointers are chosen to represent initial cluster centers, all data points are assigned to the nearest one, the mean value of the points in each cluster is computed to form its new cluster centre and iteration continues until there are no changes in the clusters. The K-means algorithms iterates over the whole dataset until convergence is reached.

### 3.2.1 Methodology used

The K-means algorithm is a most well-known and commonly used partitioning method. It takes the input parameter, K, and partitions a set of N objects into K clusters so that the resulting intra-cluster similarity is high but the inter cluster similarity is low. Cluster similarity is measured in regard to the mean value of the objects in a cluster [8]. The working of algorithm is like it randomly selects the K objects, each of which initially represents cluster mean or center. For each of the remaining objects, an object is assigned to the cluster to which it is the most similar, based on the distance between the objects and the cluster mean. It then computes the new mean for each cluster. This process iterates until the criterion function converges.

An important step in most clustering is to select a distance measure, which will determine how the similarity of the two elements is calculated. This will influenced the shape of the clusters, as some elements may be close to one another according to one distance and farther away according to one another. Another important distinction is whether the clustering uses symmetric or asymmetric distances [8]. Many of the distance function have the property that distances are symmetric. Here, we are using the binary variables. A binary variable has two states 0 or 1, where 0 means that variable is absent and 1 means that is present. In this study, we use the partitioning method K-means algorithm, where each cluster is represented by the mean value of the objects in the cluster. In this partitioning method, the database has N objects or data tuples, it constructs K partitions of the data, where each partition represents a cluster and it classifies the data into K groups. Each group contains at least one object and each object must belong to exactly one group. The clustering results obtained by us are shown under:

Clustered Instances      LD = 0 (No)     -     94 Nos.  -  75.20 %
Clustered Instances      LD = 1 (Yes)     -     31 Nos.  -  24.80 %

The clustering history and the cluster visulizer, indicating LD = Y and LD = N are as shown in Table 3 and in Figure 2 respectively below.

Table 3. Clustering history

| Sl. No | Attributes | Full Data (125) | LD = 0 (No) | LD = 1 (Yes) |
|--------|-----------|-----------------|-------------|--------------|
| 1 | DR | 50.224 | 48.787 | 54.581 |
| 2 | DS | 6.408 | 6.383 | 6.484 |
| 3 | DH | 30.744 | 30.596 | 31.194 |
| 4 | DWE | 85.552 | 86.351 | 83.129 |
| 5 | DBA | 85.008 | 86.511 | 80.452 |
| 6 | DHA | 79.712 | 81.011 | 75.774 |
| 7 | DA | 67.312 | 67.734 | 66.032 |
| 8 | ED | 83.960 | 85.319 | 79.839 |
| 9 | DM | 85.240 | 86.245 | 82.194 |
| 10 | LM | 82.176 | 80.766 | 86.452 |
| 11 | DSS | 84.632 | 81.436 | 94.323 |
| 12 | DNS | 81.344 | 78.787 | 89.097 |
| 13 | DLL | 83.400 | 81.457 | 89.290 |
| 14 | DLS | 86.200 | 82.521 | 97.355 |
| 15 | STL | 83.632 | 83.117 | 85.194 |
| 16 | RG | 85.248 | 84.692 | 86.936 |
| No. of iterations | | | | 2 |
| Within cluster sum of squared errors | | | | 86.105 |
| Missing values globally replaced with mean/mode | | | | |

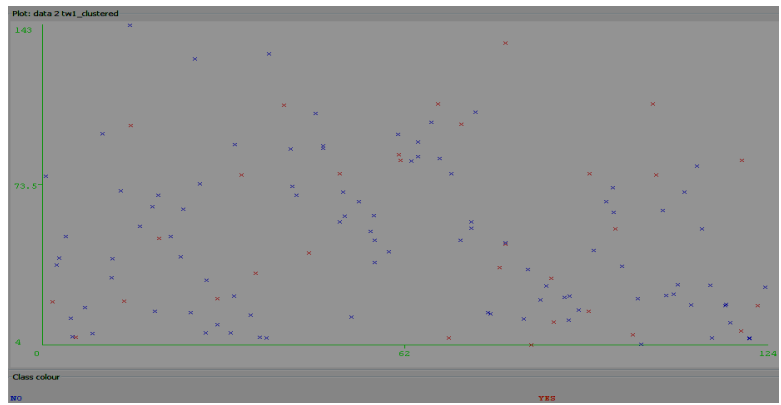

Figure 2. Cluster visulizer

# 4. RESULT ANALYSIS

In this study, we are used 125 real data sets with 16 attributes most of which takes binary values for the LD classifications. J48 algorithms are found very suitable for handling missing values and the key symptoms of LD can easily be predicted. The decision tree is very user friendly architecture compared to other classification methods. J48 decision tree is better in terms of efficiency and complexity. From this study, we have obtained that; decision tree correctly classified 77.6 % of instances. The key symptoms of LD are determined by using the attribute selection method in decision tree. By using decision tree, simple and very effective rules can be formed for LD prediction. It is also found that in case of inconsistent data, decision tree provides no solution. The accuracy of decision making can also be improved by applying the rules formulated from the tree. On comparing with our other recent studies focused on RST, SVM & MLP, Decision tree is found best in terms of efficiency and complexity,

From the study, it is also found that clustering, as one of the first step in data mining analysis, identifies groups of related records that can be used as a starting points for exploring further relationship. This technique supports the development of classification models of LD such as LD-Yes or LD-No and also formed the attribute clusters present in LD-Yes and LD-No. From the results obtained from clustering classification, we found the importance of attributes in predicting LD. In clustering also we have used the same 125 real data sets with 16 attributes.

## 5. COMPARISON OF RESULTS

In this study, we are used the algorithms J48 and K-means for prediction of LD in children. The results obtained from this study are compared with the output of a similar study conducted by us using Rough Set Theory (RST) with LEM1 algorithm. From these, we have seen that, the rules generated based on decision tree is more powerful than those of rough set theory. From the comparison of results, we have also noticed that, decision tree algorithm, J48, has a number of advantages over RST with LEM1 algorithm for solving the similar nature of problems. For large data sets, there may be chances of some incomplete data or attributes. In data mining concept, it is difficult to mine rules from these incomplete data sets. In decision tree, the rules formulated will never influenced by any such incomplete datasets or attributes. Hence, LD can easily be predicted by using the methods adopted by us. The other benefit of decision tree concept is that it leads to significant advantages in many areas including knowledge discovery, machine learning and expert system. Also it may act as a knowledge discovery tool in uncovering rules for the diagnosis of LD affected children. The importance of this study is that, using a decision tree we can easily predict the key attributes (signs and symptoms) of LD and can predict whether a child has LD or not. For very large data set, the number of clusters can easily be identified using clustering method.

Obviously, as the school class strength is 40 or so, the manpower and time needed for the assessment of LD in children is very high. But using the techniques adopted by us, we can easily predict the learning disability of any child. Decision tree approach shows, its capability in discovering knowledge behind the LD identification procedure. The main contribution of this study is the selection of the best attributes that has the capability to predict LD. In best of our knowledge, none of the rules discovered in this type of study, so far, have minimum number of attributes, as we obtained, for prediction of LD. The discovered rules also prove its potential in correct identification of children with learning disabilities.

## 6. CONCLUSION AND FUTURE RESEARCH

In this paper, we consider an approach to handle learning disability database to predict frequent signs and symptoms of the learning disability in school age children. This study mainly focuses on two classification techniques, decision tree and clustering, because accuracy of decision-making can be improved by applying these methods. This study has been carried out on 125 real data sets with most of the attributes takes binary values and more work need to be carried out on quantitative data as that is an important part of any data set. In future, more research is required to apply the same approach for large data set consisting of all relevant attributes. This study is a true comparison of the proposed approach by applying it to large datasets and analyzing the completeness and effectiveness of the generated rules.

J48 decision tree application on discrete data and twofold test shows that it is better than RST in terms of efficiency and complexity. J48 decision tree has to be applied on continuous or categorical data. Noise effects and their elimination have to be studied. The results from the experiments on these small datasets suggests that J48 decision tree can serve as a model for classification as it generates simpler rules and remove irrelevant attributes at a stage prior to tree induction. By using clustering method, the number of clusters can easily be identified in case of very large data sets.

In this paper, we are considering an approach to handle learning disability database and predicting the learning disability in school age children. Our future research work focuses on, fuzzy sets, to predict the percentage of LD, in each child, thus to explore the possibilities of getting more accurate and effective results in prediction of LD.

Conference on Advanced Computing, ICAC 2009, MacMillion Publishers India Ltd., NYC, ISBN 10:0230-63915-1, ISBN 13:978-0230-63915-7, pp. 202–207

**Julie M. David** born in 1976 received the Masters degree in Computer Applications (MCA) from Bharathiyar University, Coimbatore, India and M.Phil degree in Computer Science from Vinayaka Missions University, Salem, India in 2000 and 2008 respectively. She is currently pursuing Ph. D in the area of Data Mining at Cochin University of Science and Technology, Cochin, India. During 2000-2007 she was with Mahatma Gandhi University, Kottayam, India as a Lecturer in the Department of Computer Applications. She is now with MES College, Aluva, Cochin, India as an Asst. Professor in the Department of Computer Applications. She has published several papers in international and national conference proceedings. Her research interest includes Data Mining, Artificial Intelligence and Machine Learning. She is a member of International Association of Engineers and a reviewer of Elsevier Knowledge Based Systems.

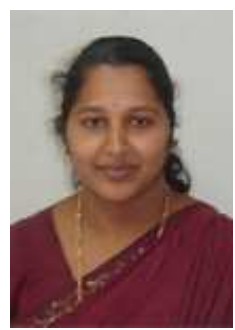

**Dr. Kannan Balakrishnan** born in 1960 received the M.Sc and M. Phil degrees in Mathematics from University of Kerala, India, M. Tech degree in Computer and Information Science from Cochin University of Science & Technology, Cochin, India and Ph. D in Futures Studies from University of Kerala, India in 1982, 1983, 1988 and 2006 respectively. He is currently working with Cochin University of Science & Technology, Cochin, India, as an Associate Professor (Reader) in the Department of Computer Applications. He has visited Netherlands as part of a MHRD project on Computer Networks. Also he visited Slovenia as the co-investigator of Indo-Slovenian joint research project by Department of Science and Technology, Government of India. He has published several papers in international journals and national and international conference proceedings. His present areas of interest are Graph Algorithms, Intelligent Systems, Image Processing, CBIR and Machine Translation. He is a reviewer of American Mathematical Reviews. He is a recognized Research Guide in the Faculties of Technology and Science in the Cochin University of Science and Technology, Cochin, India. He has served in many academic bodies of various universities in Kerala, India. Also currently he is a member of the Board of Studies of Cochin, Calicut and Kannur Universities in India. He is also a member of MIR labs India.

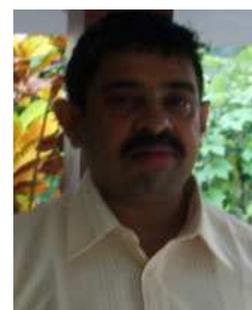